\documentclass{article}

\usepackage{hyperref}       %
\usepackage{url}            %

 \usepackage{amsmath,amssymb,amsbsy,amsthm,amsfonts,natbib}

\let\epsilon\varepsilon
\let\phi\varphi

\def\X{{\cal X}}

\def\N{\mathbb N}

\def\C{\mathcal C}

\def\E{{\bf E}}

\def\-as{\text{-a.s.}}

 \newtheorem{theorem}{Theorem}
 \newtheorem*{theorem*}{Theorem}

 \theoremstyle{remark}

\title{Finite-time optimality of Bayesian  predictors}

\date{}

\author{Daniil Ryabko}

\begin{document}

\maketitle

\begin{abstract}
The problem of sequential probability forecasting is considered in the most general setting: a model set $\C$ is given, and it is required to predict as well as possible if any of the  measures (environments)  in $\C$ is chosen to generate the data. No  assumptions whatsoever are made on the model class $\C$, in particular, no independence or mixing assumptions; $
C$ may not be measurable;  there may be no predictor whose loss is sublinear, etc.
It is shown that the cumulative loss of any possible predictor can be matched by that of a Bayesian predictor whose prior is discrete and is concentrated on $\C$, up to an additive  term of order $\log n$, where $n$ is the time step. The bound holds for every $n$ and every measure in $\C$.  This is the first non-asymptotic result of this kind. In addition, a non-matching lower bound is established: it goes to infinity with $n$ but may do so arbitrarily slow.
\end{abstract}

\section{Introduction}
Choosing a model is a hard problem. Its  solutions are often driven by the ease of finding an algorithm rather than by the adequacy of the model to the task at hand. In the context of the prediction problem, at the very least it is  typically assumed that a predictor whose loss is sublinear exists. 
Even under this assumption, there are no generic methods for constructing a predictor given only a model set $\C$. This applies not only to the prediction problem, but more generally. 
One generic method for constructing a learning algorithm is the Bayesian one: choose a prior over the model class and predict according to the posterior distribution given the data. However, there are classical results that show that some, or even, in some sense, most of the priors result in an inconsistent method \citep{Freedman:63,Freedman:65,Diaconis:86}.  Therefore, the question arises: is it possible to show that the Bayesian  with at least {\em some} prior will be  optimal?  In the asymptotic sense, this question was answered in the positive by \citet{Ryabko:10pq3+, Ryabko:17pq5}
(first  under the assumption that the best achievable asymptotic  average error is 0, and then without this assumption). Thus, the  smallest asymptotic average error  is achieved by a Bayesian predictor with some prior. However, this leaves open the question of what happens before infinity, allowing for the possibility that, for finite $n$,  every Bayesian predictor is grossly suboptimal. 

Here we resolve this doubt,  and show that, for any model set $\C$, there is a prior over this set, such that the Bayesian predictor with this prior has optimal  cumulative error up to an additive term of order $ \log n$   for every  time step $n$ for every measure $\mu$ in $\C$ (not just with prior average). This means that, the regret of being   Bayesian, with some prior, on time-average is at most   $O(\log n/n)$. This is generally considered rather small; in particular, already the fact that there is no multiplicative factor may be remarkable, and already this is new for the case of  $o(n)$ cumulative loss.  We also establish a lower bound, though the gap with respect to the upper bound is relatively large: the lower bound on the cumulative regret of being a Bayesian goes to infinity but may do so arbitrarily slow.

{\bf\noindent Setup.}
A bit more formally, the problem is that of sequential probability forecasting in the following setting. Given a  sequence $x_1,\dots,x_n$ of observations $x_i\in\X$, where $\X$ is a finite set, it is required to predict
 the probabilities  of observing $x_{n+1}=x$ for each $x\in\X$, before $x_{n+1}$ is revealed,
after which the process continues sequentially, $n=1,2,\dots$. The problem is considered in full generality; in particular, outcomes may exhibit arbitrary dependence.
What is given is a set $\C$ of measures over the space of all one-way infinite sequences. It is assumed that one of the measures in $\C$, say $\mu$,  is chosen to generate the data, and it is required to construct a predictor whose error is as small as possible for every $\mu\in\C$.
The error is measured in terms of the expected (with respect to the unknown $\mu$ measure that generates the data) cumulative (over $n$ time steps) KL divergence (log loss) $L_n$:
\begin{equation*}%
  L_n(\mu,\rho):= 
   \E_\mu
  \sum_{t=1}^n  \sum_{a\in\X} \mu(x_{t}=a|x_{1..t-1}) \log \frac{\mu(x_{t}=a|x_{1..t-1})}{\rho(x_{t}=a|x_{1..t-1})}.
\end{equation*}

{\bf\noindent Motivation.}
 This and related problems arise in a variety of applications, where the data may be financial, such as a sequence of stock prices; human-generated, such as a written text or a behavioural sequence; biological (DNA sequences); physical measurements and so on. 
In many of these applications very little, if anything, is known about the process that generates the data, and therefore it is hard to come up with reasonable assumptions. Moreover, achieving 0 asymptotic average error is often hopeless. For example, one can never hope to learn to predict accurately the probabilities of the stock market prices, even on long-term average; nor the probability distribution of the next letter of a human-written or other natural text, a problem that is directly  linked to compressing such texts. This prompts a  consideration of very general classes of environments $\C$, that would allow for some learning yet would also encompass as much as possible all the natural environments one tries to model. 

One way to come up with such sets is considering changing environments. For instance, the data sequence may have a number of change points, such that between each two consecutive change points the sequence is generated by a relatively simple measure (e.g., i.i.d.\ or Markov), but the sequence of change points is essentially arbitrary, with the only constraint being a one on the frequency of changes. Another way may be to consider arbitrary additive trends: again, take a sequence generated by a measure from a relatively simple set and sum it up with another, which may be arbitrary except for a constraint on how fast it changes.

These are just some of the ways of constructing  large sets of measures $\C$. These methods  do not come close to fully addressing the challenges arising in applications just mentioned. Here we do not concentrate on any particular example, but rather attempt to tackle the  problem in its full generality.  %

{\noindent\bf The main result.} Take any set $\C$ of measures and an arbitrary predictor $\rho$. We show that there exists a Bayesian predictor, $\nu$, such that its excess loss with respect to $\rho$ is at most logarithmic: 
$$
 L_n(\mu,\nu)\le L_n(\mu,\rho)+ 8\log n + O(\log\log n)
$$ for every $\mu\in\C$. Moreover, the prior of the Bayesian predictor can always be chosen to be discrete, that is, $$\nu=\sum_{i=1}^\infty w_i\mu_i,$$ where $\mu_i\in\C$, and $w_i$ are real weights. This in particular allows us to consider sets $\C$ which may not be measurable.  The constants in the $O(\cdot)$ term are small and are given explicitly; apart from absolute constants, there is also a linear dependence on the size of the alphabet $|\X|$.  

This is a theoretical result. More steps remain to be made before real applications can be addressed. Perhaps the most important one is finding  a general method for constructing the optimal prior whose existence is established in this work. 

In addition, a lower bound is established, showing that  there exists a set of measures $\C$ and a measure $\rho$, such that for every Bayesian predictor $\nu$ whose prior is concentrated on $\C$, there exists a function $\theta(n)\to\infty$, there exist infinitely many time steps $n_i$ and measures $\mu_i\in\C$ such that $$L_{n_i}(\mu_i,\nu)\ge L_{n}(\mu_i,\rho) + \theta(n_i)$$ for all $i\in\N$. 

Thus, there is an order-$\log n$ gap between the upper and the lower bounds.

{\noindent\bf Related work.} Apart from the previously mentioned results on which this work builds, one should mention  an alternative general approach to prediction, namely, prediction with expert advise   \cite{Cesa:06}.  Here it is assumed that the sequence one tries to predict is completely arbitrary, but, instead, one is given a set of predictors (or experts) $\C$ to compete with.  The relations between this setting and the one considered here have been analysed by \citet{Ryabko:11pq4+}. What is important to note is that for this problem in its full generality there is so far no generic method for constructing predictors to compete with an arbitrary set of experts $\C$. In particular, \citet{Ryabko:16pqnot} shows that there are sets $\C$ such that every Bayesian predictor has suboptimal asymptotic average regret. Note that such sets must necessarily be large (in particular, uncountable), while most of the work on expert advise concentrates on finite sets of experts $\C$ or else on sets of experts satisfying some very specific properties. It remains open to find which properties of sets of experts are necessary and sufficient for any general algorithm (Bayesian or not) to be optimal.

\section{Setup}\label{s:pre}
Let $\X$ be a finite set (the alphabet), and let 
\begin{equation}\label{eq:M}
 M:=\log |\X|.
\end{equation}
  The notation $x_{1..n}$ is used for $x_1,\dots,x_n$. 
The symbol   $\E_\mu$ denotes the expectation with respect to a measure $\mu$.
 We consider (probability) measures on $(\X^\infty,\mathcal F)$, where $\mathcal F$
is the usual Borel sigma-field. 

In general, a {\em Bayesian} predictor $\nu$ over a set $\C$ is a measure $\int_\C\alpha d W(\alpha)$ where $W$ is a measure over the set of all measures on $(\X^\infty, \mathcal F)$, the latter being assumed endowed with the structure of a probability space %
\cite{Gray:88}.
However, in this paper we shall only be dealing with Bayesian predictors with discrete priors, that is, with predictors of the form $\sum_{i=1}^\infty w_i\mu_i$, where $(w_i)_{i\in\N}$ are reals (that play the role of the distribution $W$ above), and $\mu_i\in\C$, $i\in\N$. This allows us to avoid any measureability issues, in particular allowing $\C$ to be non-measurable.

For two measures $\mu$ and $\rho$ introduce the {\em expected cumulative Kullback-Leibler divergence (KL divergence)} as
\begin{multline}\label{eq:kl} 
  L_n(\mu,\rho):= 
   \E_\mu
  \sum_{t=1}^{n-1}  \sum_{a\in\X} \mu(x_{t}=a|x_{1..t-1}) \log \frac{\mu(x_{t}=a|x_{1..t-1})}{\rho(x_{t}=a|x_{1..t-1})}
 \\=
  \sum_{x_{1..n}\in\X^n}\mu(x_{1..n}) \log \frac{\mu(x_{1..n})}{\rho(x_{1..n})}.
\end{multline}
In words, we take the $\mu$-expected (over data) cumulative (over time) KL divergence between $\mu$- and $\rho$-conditional (on the past data) 
probability distributions of the next outcome; and this gives simply the $\mu$-expected log-ratio of the likelihoods.
Here $\mu$ will be interpreted as the distribution generating the data.

\section{Main result}\label{s:main}
The main result shows that  the performance of any predictor can be matched by that of a Bayesian predictor with some prior, up to an additive $\log n$ term.

\begin{theorem}\label{th:main}Let $\C$ be any set of probability measures on $(\X^\infty,\mathcal F)$,
and let $\rho$ be another probability measure on this space, considered as a predictor. Then there is a  discrete Bayesian predictor $\nu$, 
that is, a predictor of the form $\sum_{k\in\N} w_k \mu_k$ where $\mu_k\in\C$ and $w_k\in[0,1]$, such that for every  $\mu\in\C$ we have
\begin{equation}\label{eq:thm}
 L_n(\mu,\nu)- L_n(\mu,\rho) \le  8 \log n + O(\log\log n),
\end{equation}
 where the constants in $O(\cdot)$ are small and are given in \eqref{eq:close}  using the notation defined in \eqref{eq:M}, \eqref{eq:w}, \eqref{eq:B} and~\eqref{eq:k}. The dependence on the alphabet size, $M$, is linear ($M\log\log n$) and the rest of the constants are universal.
\end{theorem}

The proof (which follows below) uses the construction from \cite{Ryabko:17pq5} with a refined and added analysis that allows for rates extraction.
The main ideas of the proof are as follows.
First of all,  a separate predictor is constructed to work on time  steps $1$ to $n$ for each $n$; these predictors are later summed up with weights to obtain the final predictor. Before going any further, note that constructing a predictor for each $n$ must be done without forgetting the rest of the time indices $n$: in fact, taking a predictor that is minimax optimal for each $n$ and summing these predictors up (with weights) for all $n\in\N$ may result in the worst possible predictor overall, and in particular, a one much worse than the predictor $\rho$ given. An example of this behaviour is given in the proof of Theorem~\ref{th:lb} (the lower bound). Thus, the measure $\rho$ is used in an essential way when constructing a predictor for each of the time steps $n$.  For each $n$, we consider a covering of the set $\X^n$ with subsets, each of which is associated with a measure $\mu$ from $\C$. These latter measures are then those the prior is concentrated on (that is, they are summed up with weights). The covering is constructed as follows. The log-ratio function $\log\frac{\mu(x_{1..n})}{\rho(x_{1..n})}$, where $\rho$ is the predictor whose performance we are trying to match, is approximated with a step function for each $\mu$, and for each size of the step. The cells of the resulting partition are then ordered with respect to their $\rho$ probability. The main part of the proof is then to show that not too many cells are needed to cover the set $\X^n$ this way up to a small probability. Quantifying the ``not too many'' and ``small'' parts results in the final bound.

\begin{proof}[of Theorem~\ref{th:main}.]
Define the weights $w_k$ as follows: $w_1:=1/2$, and, for $k>1$
\begin{equation}\label{eq:w} 
w_k:=w/k\log^2k,
\end{equation}
 where $w$ is the normalizer such that $\sum_{k\in\N}w_k=1$.
%Recall  the notation $M:=\log|\X|$.
Replacing $\rho$ with $1/2(\rho+\delta)$ if necessary, where $\delta$ is the i.i.d.\ probability measure with equal probabilities of outcomes, i.e.\ $\delta(x_{1..n})=M^{-1/n}$ for all $n\in\N, x_{1..n}\in\X^n$,
we shall assume, without loss of generality,
\begin{equation}\label{eq:boundedness}
  -\log\rho(x_{1..n})\le nM+1  \text{ for all $n\in\N$
and $x_{1..n}\in\X^n$}.                      
\end{equation}
The replacement is without loss of generality as it adds at most $1$ to the final bound (to be accounted for).
% 
% 
% When speaking about measures $\nu$ that we construct as countable convex combinations of measures in $\C$
% we will assume w.l.o.g.\  
% (the value of the constant will not be important, but one can take $const=1$);
 Thus, in particular, 
\begin{equation}\label{eq:boundedness33}
L_n(\mu,\rho)\le nM+1\text{ for all $\mu$}. 
 \end{equation}

The first part of the proof is the following covering construction. % from~\citep{Ryabko:17pq5}, which follows.

For each $\mu\in\C$, $n\in\N$ define the sets 
\begin{equation}\label{eq:tt}
T_{\mu}^n:=\left\{x_{1..n}\in \X^n:  \frac{\mu(x_{1..n})}{\rho(x_{1..n})}\ge {1\over n}\right\}.
\end{equation} 
From Markov inequality, we obtain 
\begin{equation}\label{eq:tm}
\mu(\X^n\backslash T_{\mu}^n)\le 1/n.
\end{equation}

For each $k>1$ let $U_k$ be the partition of  $[-\frac{\log n}{n},M+{1\over n}]$ into   $k$ intervals defined as follows.
 $U_k:=\{u_k^i:i=1..k\}$, where
\begin{equation}\label{eq:cover}
 u^i_k=\left\{ \begin{array}{ll}
                \left[-\frac{\log n}{n},{iM\over k}\right] & i=1, \\
 		\left(\frac{(i-1)M}{k},\frac{iM}{k}\right] & 1<i<k, \\
 		\left(\frac{(i-1)M}{k},M+\frac{1}{n}\right] & i=k.
               \end{array}
  \right.
\end{equation}
Thus, $U_k$ is a partition of $[0,M]$ into $k$ equal intervals but for  some  padding that we added to the leftmost and the rightmost intervals:
on the left we added $[-\frac{\log n}{n},0)$ and on the right $(M,M+1/n]$.
% $u^1_k:=$,   $u^i_k:=$ for $1<i<k$ and  $u^k_k:=$. 

For each $\mu\in\C$, $n,k>1$, $i=1..k$ define the sets 
\begin{equation}\label{eq:t}
T_{\mu,k,i}^n:=\left\{x_{1..n}\in \X^n: {1\over n}\log \frac{\mu(x_{1..n})}{\rho(x_{1..n})}\in u_k^i\right\}.
\end{equation} 
Observe that,  for every  $\mu\in\C,$ $k,n>1$, these sets  constitute a partition of % $\X^n$
$T_{\mu}^n$ into $k$ disjoint sets: indeed, on the left we have ${1\over n}\log \frac{\mu(x_{1..n})}{\rho(x_{1..n})}\ge -{1\over n}\log n$ 
 by definition~\eqref{eq:tt}
of $T_{\mu}^n$, and on the right we have ${1\over n}\log \frac{\mu(x_{1..n})}{\rho(x_{1..n})}\le M+1/n$ from~\eqref{eq:boundedness}.
In particular, from this definition,    %Note that by construction~\eqref{eq:t} 
for all $x_{1..n}\in T^n_{\mu,k,i}$ we have
 \begin{equation}\label{eq:cops1}
 % 2^{{(i-1)M\over k}n}\rho(x_{1..n})\le 
 \mu(x_{1..n}) \le 2^{{iM\over k}n+1}\rho(x_{1..n}). %%% +1 from the case i=1 (say this?)
 \end{equation}
%{\em Step 2n: a countable cover, time $n$.}

%Fix an $n\in\N$.
 For every $n,k\in\N$ and $i\in\{1..k\}$ consider the following construction. 
 Define 
$$m_{1}:=\max_{\mu\in\C}\rho(T_{\mu,k,i}^n)$$ (since $\X^n$ are finite all suprema are reached). 
 Find any $\mu_{1}$ such that $\rho(T_{\mu_1,k,i}^n)=m_{1}$ and let
$T_{1}:=T^n_{\mu_1,k,i}$. For $l>1$, let 
$$m_{l}:=\max_{\mu\in\C}\rho(T_{\mu,k,i}^n\backslash T_{l-1}).$$
 If $m_l>0$, let $\mu_l$ be any $\mu\in\C$ such 
that $\rho(T^n_{\mu_l,k,i}\backslash T_{l-1})=m_l$, and let $T_l:=T_{l-1}\cup T^n_{\mu_l,k,i}$; otherwise let $T_l:=T_{l-1}$ and $\mu_l:=\mu_{l-1}$. 
%Note that, since $\mu\in\C$, we must have $T^n_{\mu,k,i}=\cup_{l\in\N}T^n_{\mu_l,k,i}$. Therefore, 
Note that,  for each $x_{1..n}\in T_{l}$ there is $l'\le l$ such that $x_{1..n}\in T^n_{\mu_{l'},k,i}$ and thus from~\eqref{eq:t} we get      %(similar to~\eqref{eq:cops1})
 \begin{equation}\label{eq:cops2}
  2^{{(i-1)M\over k}n-\log n}\rho(x_{1..n})\le \mu_{l'}(x_{1..n}).  %%%%\log n from the case i=1
   % \le 2^{{iM\over k}n}\rho(x_{1..n}).
 \end{equation}
 Finally, define 
\begin{equation}\label{eq:nun}
\nu_{n,k,i}:=\sum_{l=1}^{\infty} w_l\mu_l.
\end{equation}
(Notice that 
for every $n,k,i$ there is only a finite number of positive $m_l$,
since the set $\X^n$ is finite; thus the sum in the last definition is effectively finite.)
% We will show that the set $\{\nu_{n,k,i}:n,k\in\N, i=1..k\}$ is the countable set (sequence) which we are looking for to establish~\eqref{eq:nuro}.
%Thus, we shall define the predictor $\nu$ as 
Finally, define the predictor $\nu$ as 
\begin{equation}\label{eq:nu}
\nu:={1\over2}\sum_{n,k\in\N}w_nw_k{1\over k}\sum_{ i=1}^k\nu_{n,k,i} +{1\over2}r,
\end{equation}
where $r$ is a regularizer defined so as to have for each $\mu'\in\C$ and $n\in\N$
\begin{equation}\label{eq:boundedness2}
  \log\frac{\mu'(x_{1..n})}{\nu(x_{1..n})}\le nM - \log w_n +1  \text{\ \ for all  $x_{1..n}\in\X^n$};                      
\end{equation}
this and the stronger statement  \eqref{eq:boundedness} for $\nu$  can be obtained analogously to the latter inequality in the case the i.i.d.\ measure $\delta$ is in $\C$; otherwise (since we need to define $\nu$ as a combination of probability measures from $\C$ only), $r$  can be defined the same way as is done in \citep[Step~{\em r} of the proof of Theorem~5]{Ryabko:10pq3+}; for the sake of completeness, this argument is given in the end of this proof.

Next, let us show that the measure $\nu$ is the predictor whose existence is claimed in the statement. %~\eqref{eq:nuro} holds for $\nu$ so defined.

Introduce the notation 
$$
L_n|_A(\mu,\nu):=\sum_{x_{1..n}\in A}\mu(x_{1..n})\log\frac{\mu(x_{1..n})}{\rho(x_{1..n})};
$$
with this notation, for any set $A\subset\X^n$ we have
$$
L_n(\mu,\nu)= L_n|_A(\mu,\nu)+L_n|_{\X^n\setminus A} (\mu,\nu).
$$

First we want to show that, for each $\mu\in\C$, for each fixed $k,i$,  the sets $T^n_{\mu,k,i}$ are covered by sufficiently few 
sets  $T_l$, where ``sufficiently few'' is, in fact, exponentially many with the right exponent. 
By definition, for each $n,i,k$ the sets $T_l\backslash T_{l-1}$ are disjoint (for different $l$) and have non-increasing (with $l$) $\rho$-probability. Therefore, $\rho(T_{l+1}\backslash T_{l})\le 1/l$  for all $l\in\N$. Hence, from the definition of $T_l$, we must also have $\rho(T^n_{\mu,k,i}\backslash T_{l})\le 1/l$   for all $l\in\N$.
From the latter inequality and~\eqref{eq:cops1} we obtain 
$$
\mu(T^n_{\mu,k,i}\backslash T_{l})\le{1\over l} 2^{{iM\over k}n+1}.
$$
%Consequently, for any $a>M/k$ 
 Take $l_i:=\lceil{ kn}2^{{iM\over k}n+1}\rceil$ to obtain 
\begin{equation}\label{eq:li}
 \mu(T^n_{\mu,k,i}\backslash T_{l_i})\le {1\over kn}.
\end{equation}

Moreover, for every $i=1..k$, for each  $x_{1..n}\in T_{l_i}$,  %{\mu,k,i}$  %%% typo k--l ?
 %except possibly for a set of $\mu$-probability~${1\over k}2^{-an}$ (that is, for $x_{1..n}\in T^n_{\mu,k,i}\backslash T_{l_i}$) 
there is $l'\le l_i$ such that $x_{1..n}\in T^n_{\mu_{l'},k,i}$ and thus   the following chain holds
\begin{multline}\label{eq:fnu}
% \nu(x_{1..n})\ge w_nw_kw_i w_{2^{({i\over k}+a)n}}\mu(x_{1..n})\ge w_nw_kw_i {2^{-({i\over k}+a)n}}\mu(x_{1..n}) \ge w_nw_kw_i {2^{-(a-{1\over k})n}}\rho(x_{1..n})
 \nu(x_{1..n})\ge {1\over2}w_nw_k{1\over k} \nu_{n,k,i} \ge {1\over2}w_nw_k{1\over k} w_{kn\,2^{{iM\over k}n+1}}\mu_{l'}(x_{1..n}) 
\\ 
 \ge {w^3\over 4n^2k^3\log^2n\log^2k (\log k+\log n+1+ nMi/k)^2} 2^{-{iM\over k}n}\mu_{l'}(x_{1..n}) 
\\ 
 \ge {w^3\over 4(M+1)^2n^4k^3\log^2n\log^2k} 2^{-{iM\over k}n}\mu_{l'}(x_{1..n}) 
 \\
 \ge {w^3\over 4(M+1)^2n^5k^3\log^2n\log^2k} 2^{-{M\over k}n}\rho(x_{1..n}) 
%\\
= B_n  2^{-{M\over k}n}\rho(x_{1..n}),
\end{multline}
where the first inequality is from~\eqref{eq:nu}, the second from~\eqref{eq:nun} with  $l=l_i$, the third is 
by definition of $w_l$, the fourth uses $i\le k$ for the exponential term,  as well as  $(\log n +\log k)\le n-1$ for $n\ge3$, which will be justified by the choice of $k$ in the following~\eqref{eq:k},
  the fifth inequality   uses~\eqref{eq:cops2},  and the final equality introduces $B_n$ defined as  
\begin{equation}\label{eq:B}
 B_n:={w^3\over 4(M+1)^2n^5k^3\log^2n\log^2k}.
\end{equation}

We have
\begin{equation}\label{eq:br}
 L_n(\mu,\nu)= \left(\sum_{i=1}^kL_n|_{T_{l_i}}(\mu,\nu)\right) +  L_n|_{\X^n\setminus \cup_{i=1}^kT_{l_i} }(\mu,\nu).
% \sum_{i=1}^kL_n|_{T^n_{\mu,k,i}\setminus T_{l_i}}(\mu,\nu) + L_n|_{T^n_\mu\setminus \cup_{i=1}^k T^n_{\mu,k,i}}(\mu,\nu)  + L_n|_{\X^n\setminus T^n_\mu}(\mu,\nu) .
\end{equation}
For the first term, from~\eqref{eq:fnu} we obtain
\begin{multline}\label{eq:lmn1}
 \sum_{i=1}^kL_n|_{T_{l_i}}(\mu,\nu)
 \le  \sum_{i=1}^kL_n|_{T_{l_i}}(\mu,\rho) +Mn/k - \log B_n
 \\ = L_n(\mu,\rho)-L_n|_{\X^n\setminus\cup_{i=1}^kT_{l_i}}(\mu,\rho) +Mn/k - \log B_n.
\end{multline}
For the second term in~\eqref{eq:br}, we recall that $T^n_{\mu,k,i}$, $i=1..k$ is a partition of $T^n_\mu$, and  decompose 
\begin{equation}\label{eq:3sets}
 \X^n\setminus \cup_{i=1}^kT_{l_i}\subseteq\left(\cup_{i=1}^k(T^n_{\mu,k,i}\setminus T_{l_i})\right)\cup (\X^n\setminus T^n_\mu).  % \cup_{i=1}^k T^n_{\mu,k,i}) ;
\end{equation}
Next, using~\eqref{eq:boundedness2} and an upper-bound for the $\mu$-probability of each of the two sets in~\eqref{eq:3sets}, namely, \eqref{eq:li} and~\eqref{eq:tm}, as well as $k\ge 1$, we obtain 
\begin{equation}\label{eq:sb}
  L_n|_{\X^n\setminus \cup_{i=1}^kT_{l_i} }(\mu,\nu) \le (nM - \log w_n+1 ) {2\over n}.
\end{equation}

Returning to~\eqref{eq:lmn1}, from Jensen's inequality one can show (see, e.g.,  \cite[equation 11]{Ryabko:10pq3+}) that, for any set $A\subset\X^n$,  
$$
-L_n|_A(\mu,\rho)\le\mu(A)\log\rho(A)+1/2.
$$
Therefore, using~\eqref{eq:boundedness33}, similarly to~\eqref{eq:sb} we obtain
\begin{equation}\label{eq:sb2}
 -L_n|_{\X^n\setminus\cup_{i=1}^kT_{l_i}}(\mu,\rho) \le  (nM +1 ) {2\over n}+{1\over2}.
\end{equation}

Combining~\eqref{eq:br} with~\eqref{eq:lmn1}, ~\eqref{eq:sb} and~\eqref{eq:sb2} we derive 
\begin{equation}\label{eq:close}
 L_n(\mu,\nu)
\le L_n(\mu,\rho)+Mn/k-\log B_n +4M -{2\over n}(\log w_n - 1)+1/2;
\end{equation}
setting 
\begin{equation}\label{eq:k}
 k:=\lceil n/\log\log n\rceil 
\end{equation}
 we obtain the statement of the theorem.

It remains to come back to~\eqref{eq:boundedness2} and define the regularizer $r$ as a combination of measures from $\C$ for this inequality to hold.
For each $n\in\N$, denote
$$
A_n:=\{x_{1..n}\in \X^n: \exists\mu\in\C\ \mu(x_{1..n})\ne0\},
$$ and let, for each $x_{1..n}\in \X^n$, the probability measure $\mu_{x_{1..n}}$ be any probability measure from $\C$ such that $\mu_{x_{1..n}}(x_{1..n})\ge{1\over2}\sup_{\mu\in\C}\mu(x_{1..n})$.
Define 
$$
 r_n'%(x'_{1..n})
   :={1\over |A_n|}\sum_{x_{1..n}\in A_n}\mu_{x_{1..n}}%(x'_{1..n}),
$$ for each
%$x'_{1..n}\in A^n$,
 $n\in\N$, and let  $r:=\sum_{n\in\N}w_n r'_n$. 
For every $\mu\in\C$ we have 
$$
r(x_{1..n})\ge w_n|A_n|^{-1} \mu_{x_{1..n}}(x_{1..n})\ge{1\over2} w_n |\X|^{-n} \mu(x_{1..n})
$$ for
every $n\in\N$ and every $x_{1..n}\in A_n$,  establishing~(\ref{eq:boundedness2}).
\end{proof}

\section{Lower bound}
In this section we establish a lower bound on being a Bayesian with  the best prior. The bound leaves a significant gap with respect to the upper bound, but it  shows that the regret of using the Bayesian predictor with the {\em best} prior for the given problem  cannot be upper-bounded by a constant.
\begin{theorem}\label{th:lb}
 There exists a set of measures $\C$ and a measure $\rho$, such that for every Bayesian predictor $\nu$ whose prior is concentrated on $\C$, there exists a function $\theta(n)$ which is non-decreasing and goes to infinity with $n$, there exist infinitely many time steps $n_i$ and measures $\mu_i\in\C$ such that $L_{n_i}(\mu_i,\nu)-L_{n_i}(\mu_i,\rho)\ge \theta(n_i)$ for all $i\in\N$.
\end{theorem}
Thus, the lower bound goes to infinity with $n$ but may do so arbitrarily slow. This leaves a gap with respect to the $O(\log n)$ upper bound of Theorem~\ref{th:main}. %
 
Note also that this formulation is good enough to be the opposite of Theorem~\ref{th:main}, because the formulation of the latter is strong: Theorem~\ref{th:main} says that {\em for every $\mu$ and for every $n$} (the regret is upper bounded), so, in order to counter that, it is enough to say that {\em there exists $n$ and there exists $\mu$} (such that the regret is lower bounded);  Theorem~\ref{th:lb} is, in fact, a bit stronger, since it establishes that there are  infinitely many such $n$.  However, it does not preclude that for every $\mu$ in $\C$ the loss of the Bayesian is upper-bounded by a constant independent of $n$, while the loss of $\rho$ is linear in $n$. This is, in fact, the case in the proof. 
\begin{proof}
Let $\X:=\{0,1\}$. Let $\C$ be the set of Dirac delta measures, that is,  the measures  each of which is concentrated on a single deterministic sequence, where the sequences are all sequences that are 0 from some $n$ on. In particular, introduce 
$S_n:=\{ x_{1,2,\dots}\in\X^\infty : x_i=0\text{ for all }i>n\}$, $S:=\cup_{n\in\N}S_n$. Let $\C_n$ be the set of all measures $\mu$ such that $\mu(x)=1$ for some $x\in S_n$ and let $\C:=\cup_{n\in\N}C_n$.   

Observe that the set $\C$ is countable. It is therefore, very easy to construct a (Bayesian) predictor for this set: enumerate it in any way, say $(\mu_k)_{k\in\N}$ spans all of $\C$, fix a sequence of positive weights $w_k$ that sum to 1,  and let 
\begin{equation}\label{eq:bn}
\nu:=\sum_{k\in\N}w_k \mu_k. 
\end{equation}
Then $L_n(\mu_k,\nu)\le-\log w_k$ for all $k\in\N$. That is, for every $\mu\in\C$ the loss of $\nu$ is upper-bounded by a constant: it depends on $\mu$ but not on the time index $n$.  So, it is good for every $\mu$ for large $n$, but may be bad for some $\mu$ for (relatively) small $n$, which is what we shall exploit.

Observe that, since $\C$ is countable, every Bayesian $\nu$ with its prior over $\C$  must have, by definition, the form~\eqref{eq:bn} for some weights $w_k\in[0,1]$ and some measures $\mu_k\in\C$.  Thus, we fix any Bayesian $\nu$ in this form.

Define $\rho$ to be the Bernoulli i.i.d.\ measure with the parameter 1/2. Note that 
\begin{equation}\label{eq:rho}
L_n(\mu,\rho)=n  
\end{equation}
for every $n$. This is quite a useless predictor; its asymptotic average error is the worst possible, 1.  However, it is minimax optimal for every single time step $n$:
$$
\inf_{\rho'}\sup_{\mu\in\C} L_n(\mu,\rho') = n,  
$$
where the $\inf$ is over all possible measures. This is why $\rho$ is hard to compete with--- and, incidentally, why being minimax optimal for each $n$ separately may be useless.

For each $s\in\N$, let $W_s$ be the weight that $\nu$ spends on the measures in the sets $\C_k$ with $k<s$, and let $M_s$ be the set of these measures: 
$$
W_s:=\sum \{w_i: \exists k<s\text{ such that }\mu_i\in \C_k\},
$$
and 
$$
M_s:= \{\mu_i: \exists k<s\text{ such that }\mu_i\in \C_k\}.
$$

By construction, 
\begin{equation}\label{eq:to1}
  \lim_{s\to\infty}W_s=1.
\end{equation}
Next, for each $n\in\N$, let $U_n:=S_{n+1}\setminus S_n$ (these are all the sequences in $S_{n+1}$ with $1$ on the $n$th position). Note that $\mu(U_{n})=0$ for each $\mu\in M_n$, while $|U_n|=2^n$. 
From the latter equality, there exists $x_{1..n}\in\X^{n+1}$ and $\mu\in U_n\subset\C_{n+1}$ such that 
$$ 
 \mu(x_{1..n}=1)\text{ but }\nu(x_{1..n})\le 2^{-n}(1-W_s).
$$
This, \eqref{eq:to1} and~\eqref{eq:rho} imply the statement of the theorem.
\end{proof}

\section{Conclusion and future work}\label{s:co}
The main result, Theorem~\ref{th:main}, is the first one to show finite-time optimality of the Bayesian method for the prediction problem in full generality; or, perhaps, at this generality, for any learning problem. A number of important questions remain, both directly extending the result of this work and more general. 

{\noindent\bf Lower bounds, necessity of the $\log n$ term.}  The first question is how sharp is the result. So far, the lower bound only shows that, for every prior, the Bayesian may suffer more than constant regret.  The  question whether the $\log n$ term is necessary remains open. 
If it is, one can ask what is the best constant in front. In the proof of the main result, the constant  comes, first of all, from all the weights used in constructing the predictor, that is, from the prior. Each of the sums in \eqref{eq:nu} contributes one or two $\log n$. The outermost is perhaps (partially) removable with some version of the doubling trick: that is, instead of summing over all time steps $n\in\N$ one would only sum over some time steps, and reuse the predictors at remaining time steps. Yet, as the proof of Theorem~\ref{th:lb} shows, some regret from summing up over different time steps is unavoidable.  The rest are less clear how to optimize. Finally, one additional $\log n$ comes from the definition of the sets $T^n_\mu$ in~\eqref{eq:tm}, via the top line of~\eqref{eq:cover}. This one would be harder to remove, because the $1\over n$ term is necessary in~\eqref{eq:sb}.

 One could also ask the question of how important it is to optimize this constant. First of all, of course,  it is only important if the $\log n$ term is at all necessary. But if it is necessary, then the constant is important, because the optimal loss is of order $\log n$ in some commonly studied special cases of $\C$, such as i.i.d.\ or Markov measures. (It is worth mentioning that  the known optimal predictors in these cases \cite{Krichevsky:93} are, in fact, Bayesian.)

  Moreover, it may be worth trying to improve the bounds specifically for the case $L_n(\mu,\rho)=O(\log n)$, since in the opposite case it is not important.

{\noindent\bf Further generalizations.} Some further natural and interesting generalizations are to different (or general) loss functions, as well as to 
infinite (countable or continuous) alphabets $\X$.

However, the most important direction for further research appears to be finding a general method of constructing a prior that results in an optimal predictor for an arbitrary class of measures $\C$. 
Another  interesting question, mentioned in the introduction, is finding out under what conditions the Bayesian procedure, or indeed any other general method, is optimal for the non-realizable version of the problem; as discussed, some conditions are necessary, as shown in \cite{Ryabko:16pqnot}.

Finally, it is also interesting to find out  to what extent the obtained result can be generalized to interactive learning problems, such as bandits, or, more generally, reinforcement learning. 

\bibliographystyle{dinat}

\end{document}